\pdfoutput=1

\documentclass[11pt]{article}

\usepackage[]{ACL2023}

\usepackage{times}
\usepackage{latexsym}

\usepackage{booktabs}
\usepackage{amsmath}
\usepackage{multirow}
\usepackage{hyperref}
\usepackage{graphicx}
\usepackage{amssymb}
\usepackage{url}

\usepackage[T1]{fontenc}

\usepackage[utf8]{inputenc}

\usepackage{microtype}

\usepackage{inconsolata}

\usepackage{xcolor}
\usepackage{soul}

\title{A General-Purpose Multilingual Document Encoder}

\author{Onur Galo\u{g}lu$^1$ ~~~ Robert Litschko$^{2}$ ~~~ Goran Glava\v{s}$^{3}$ \\ \smallskip \\
$^1$Independent Researcher\\
$^2$MaiNLP, Center for Information and Language Processing (CIS), LMU Munich, Germany \\  
$^3$CAIDAS, University of Würzburg \\
}

\begin{document}
\maketitle
\begin{abstract}
Massively multilingual pretrained transformers (MMTs) have tremendously pushed the state of the art on multilingual NLP and cross-lingual transfer of NLP models in particular. While a large body of work leveraged MMTs to mine parallel data and induce bilingual document embeddings, much less effort has been devoted to training general-purpose (massively) multilingual document encoder that can be used for both supervised and unsupervised document-level tasks. In this work, we pretrain a massively multilingual document encoder as a hierarchical transformer model (HMDE) in which a shallow document transformer contextualizes sentence representations produced by a state-of-the-art pretrained multilingual sentence encoder. We leverage Wikipedia as a readily available source of comparable documents for creating training data, and train HMDE by means of a cross-lingual contrastive objective, further exploiting the category hierarchy of Wikipedia for creation of difficult negatives. We evaluate the effectiveness of HMDE in two arguably most common and prominent cross-lingual document-level tasks: (1) cross-lingual transfer for topical document classification and (2) cross-lingual document retrieval. HMDE is significantly more effective than (i) aggregations of segment-based representations and (ii) multilingual Longformer. Crucially, owing to its massively multilingual lower transformer, HMDE successfully generalizes to languages unseen in document-level pretraining. We publicly release our code and models.\footnote{\url{https://github.com/ogaloglu/pre-training-multilingual-document-encoders}}. 
\end{abstract}

\section{Introduction}
\label{sec:introduction}

Massively multilingual Transformers (MMTs) such as XLM-R \cite{conneau2020unsupervised}, and mT5 \cite{xue2021mt5} have  drastically pushed the state-of-the-art in multilingual NLP, especially for medium-resourced languages included in their pretraining, enabling effective cross-lingual transfer of task-specific NLP models from languages with plenty of training data to languages with little or no annotated task data. 
Being standard transformer-based language models, MMTs process text linearly -- as a flat sequence of tokens, which has -- in monolingual contexts -- been shown suboptimal for document-level tasks (e.g., document classification or retrieval) for two main reasons: (1) it does not correspond to the hierarchical nature of document organization -- documents are sequences of (presumably meaningfully ordered) paragraphs,  which are in turn sequences of sentences \cite{zhang-etal-2019-hibert,glavas2020two}, and (2) representing documents longer than the MMTs maximal input length requires either document trimming, which leads to loss of potentially task-relevant information, or segmentation, which leading to context fragmentation \cite{ding-etal-2021-ernie}.   

A number of models that produce document-level representations have been proposed, albeit predominantly in the monolingual (English) realm, with two prominent lines of work. \textbf{(1)} Hierarchical encoders \cite{pappas2017multilingual,pappagari2019hierarchical,zhang-etal-2019-hibert,yang2020beyond,glavas2020two,chalkidis2022exploration} typically contextualize sentence-level representations with additional document-level parameters (e.g., an additional, document-level transformer). These document-level parameters of the encoder, added on top of a pretrained language model like BERT \cite{devlin2019bert}, are typically trained on large task-specific datasets, ranging from document classification \cite{pappagari2019hierarchical} to summarization \cite{zhang-etal-2019-hibert} and segmentation \cite{glavas2020two}. Task-specific training of document-level parameters impedes the transfer of such encoders to other tasks. \textbf{(2)} Sparse attention models \cite{child2019generating,zaheer2020big,beltagy2020longformer,tay2020sparse} modify the attention mechanism in order to reduce its computational complexity and consequently be able to encode longer texts. 
Although flat long-text encoders do not model the hierarchical nature of documents, they allow for flat encoding of substantially longer documents.           

In this work, we demonstrate the benefits of hierarchical document representations in multilingual context. We propose to train a hierarchical transformer model (HMDE), coupling (i) a pretrained multilingual sentence encoder as a lower encoder with (ii) an upper transformer that contextualizes sentence representations against each other and from which we derive document representations. Unlike in monolingual setup, where task-specific data is commonly used to train the parameters of the upper transformer \cite{zhang-etal-2019-hibert,glavas2020two}, we exploit the fact that in the multilingual context one can leverage cross-lingual document alignments to guide the \textit{pretraining} of the document encoder, i.e., its upper transformer. To this end, we leverage Wikipedia as a readily available source of quasi-parallel documents, and additionally exploit its hierarchy of categories to create hard negative examples for our contrastive pretraining objective. 

We evaluate HMDE in two arguably most prominent (cross-lingual) document-level tasks: (1) cross-lingual transfer for document classification (XLDC) and (2) cross-lingual document retrieval (CLIR). For XLDC, as a supervised task, we fine-tune HMDE on English task-specific data; in CLIR, in contrast, we leverage HDME in an unsupervised fashion, using it to produce static document embeddings (and its lower transformer to produce query embeddings). HDME exhibits performance superior to that of competitive models -- MMTs with sliding window and multilingual Longformer \cite{yu2021cross,Sagen1545786}. Crucially, HMDE generalizes well to languages unseen in its document-level pretraining. Our further analyses offer additional insights: (i) that it is important to allow updates from document-level training to propagate to the sentence-level encoder (i.e., not to freeze the parameters of the pretrained sentence encoder) and (ii) that the size of the document-level pretraining corpora matters more than its linguistic diversity (i.e., number of languages it encompasses).

\section{Hierarchical Multilingual Encoder}

\begin{figure}[t!]
    \centering
    \includegraphics[scale=0.65]{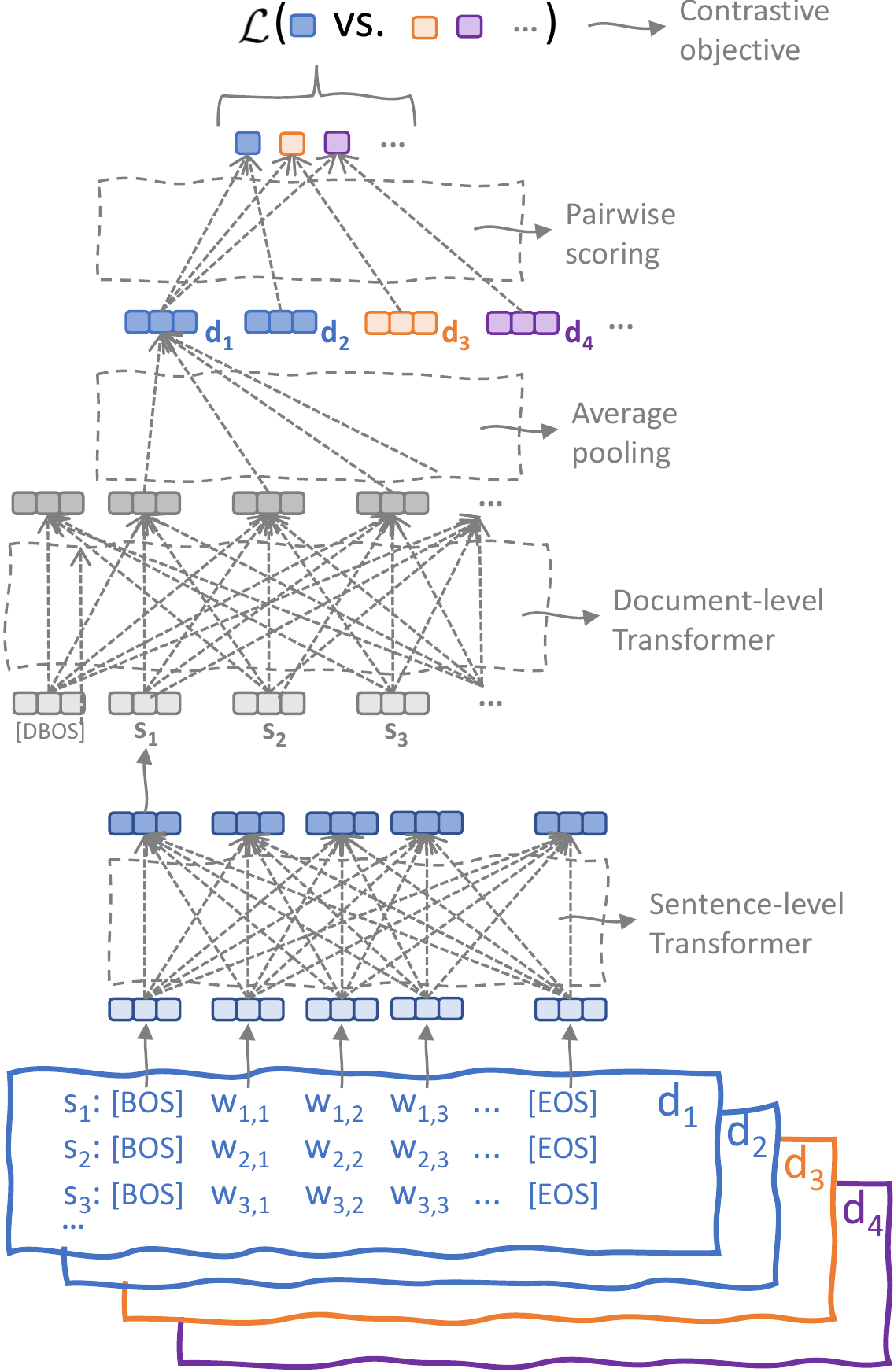}
    \caption{Illustration of HDME: hierarchical transformer architecture coupled with a cross-lingual contrastive objective. Document colors indicate the Wikipedia concepts: $d_1$ and $d_2$ are the pages of the same concept (e.g., New York) in two different languages, $L_1$ and $L_2$; documents $d_3$ and $d_4$ are pages of other concepts in $L_1$. The pair ($d_1$, $d_2$) is a positive pair (i.e., same concept) for the contrastive training objective and pairs ($d_1$, $d_3$) and ($d_1$, $d_4$) are corresponding negative pairs (i.e., different concepts).}
    \label{fig:model}
\end{figure}

The HMDE architecture, illustrated in Figure \ref{fig:model}, is similar to that of hierarchical document encoders trained monolingually in task-specific training (e.g., \cite{glavas2020two}): a sentence-level (lower) encoder produces sentence embeddings from tokens, whereas the document-level (upper) transformer yields document representation from a sequence of sentence embeddings. We initialize the lower transformer with the pretrained weights of a multilingual sentence encoder \cite{feng2022language}, and train the whole model via a bi-encoder configuration (also known as Siamese architecture) -- where we compute a similarity score between representations of two documents produced independently with HDME -- using a cross-lingual contrastive objective with both in-batch and hard negatives \cite{oord2018representation}.           

\subsection{Hierarchical Encoding}

The role of the sentence-level (lower) transformer is to produce sentence representations from sequences of tokens. Because of this, we initialize it with the pretrained weights (including subword embeddings) of LaBSE \cite{feng2022language}, a state-of-the-art multilingual sentence encoder.\footnote{We load LaBSE weights from HuggingFace: \url{https://huggingface.co/sentence-transformers/LaBSE}} The sentence embedding is the transformed representation of the special beginning-of-sequence (BOS) token. The sequence of sentence embeddings obtained with the sentence-level transformer is then forwarded to the document-level (upper) transformer, which mutually contextualizes them, prepended with a special document-level beginning-of-sequence token (DBOS, with a randomly initialized embedding). We derive the document representation  by average-pooling contextualized sentence embeddings (i.e., output of the last layer of the document-level transformer).\footnote{We preliminarily also experimented with the contextualized vector of the DBOS token as the document representation, but that consistently led to lower performance.}



\subsection{Multi- and Cross-Lingual Objective}
\label{sec:objective}

Our training dataset consists of Wikipedia pages written in one of $n$ languages (see \S\ref{sec:data} for details on the creation of different training datasets): let $L = {L_1, L_2, \dots, L_n}$ denote our set of training languages. In each training step, we select a batch of $N$ documents pairs, $\{{(d^{(1)}_1, d^{(1)}_2), \dots, (d^{(N)}_1, d^{(N)}_2)}\}$, where $d^{(i)}_1$ and $d^{(i)}_2$ are Wikipedia pages of the same concept but in two different languages, $L_k$ and $L_m \in L$. 
Each of the documents $d^{(i)}_1$ (i.e., first document of each pair) is additionally paired with a document $d^{(i)}_{\mathit{neg}}$ -- a document in the same language $L_k$ as $d^{(i)}_1$ and from the same Wikipedia category -- representing a \textit{hard negative} for $d^{(i)}_1$ (see \S\ref{sec:data} for details). We then compute and minimize a variant of the popular InfoNCE loss \cite{oord2018representation} that incorporates hard negatives, treating all other batch documents $d^{(j)}_2$ as in-batch (easy) negatives for $d^{(i)}_1$:   
\begin{multline}
\mathcal{L} = -\sum_{i = 1}^{N}{\biggl[\frac{1}{\tau}s(\mathbf{d}^{(i)}_1, \mathbf{d}^{(i)}_2)} \hspace{0.5em} - \\ \log\biggl( e^{s(\mathbf{d}^{(i)}_1,\,\mathbf{d}^{(i)}_{\mathit{neg}})/\tau} + \sum_{\substack{j = 1 \\ }}^{N}{e^{s(\mathbf{d}^{(i)}_1,\, \mathbf{d}^{(j)}_{2})/\tau}}\biggr) \biggr]    
\end{multline}
\noindent with $\mathbf{d} \in \mathbb{R}^h$ as the embedding of $d$, i.e., the output of the document-level transformer (and $h$ as the hidden size of upper transformer), $s(\mathbf{d}_i, \mathbf{d}_j)$ as the scoring function capturing similarity between the two document embeddings, and $\tau$ as the  hyperparameter (the so-called temperature) of the InfoNCE loss. Following common practice, we use  cosine similarity as the scoring function $s$. 

Note that the loss we compute is both multilingual and cross-lingual: documents $d^{(i)}_1$ come from any of the $|L|$ languages, and positive pairs $(d^{(i)}_1, d^{(i)}_2)$ are cross-lingual. Among the in-batch negatives, there will be cross-lingual as well as monolingual pairs (when $d^{(i)}_1$ and $d^{(j)}_2$ happen to be documents written in the same language). Our hard negatives are, by design, always monolingual pairs. While one could create cross-lingual hard negatives in the same manner (e.g., by pairing the English article \textit{``France''} with an Italian article \textit{``Svizzera''} (Switzerland) that covers another concept from the same category \textit{``Country''}), monolingual hard negatives should be \textit{harder} because the two document representations will originate from the same language-specific subspace of the embedding space of the lower (multilingual) transformer \cite{cao2020multilingual,wu2020explicit}.


\section{Experimental Setup}
\label{sec:experimental}

We first describe how we created the multilingual dataset for HMDE pretraining from Wikipedia (\S\ref{sec:data}). We then briefly describe the two evaluation tasks -- cross-lingual transfer for document classification and cross-lingual information retrieval -- and their respective datasets (\S\ref{sec:downstream}), following with the description of the baselines -- a multilingual sentence encoder with a sliding window and a multilingual Longformer \cite{yu2021cross,Sagen1545786} (\S\ref{sec:baselines}). We provide training and optimization details for all models in the Appendix \ref{sec:training}.    

\subsection{Data Creation}
\label{sec:data}

Wikipedia has been leveraged as a suitable source for mining comparable and parallel corpora for decades \cite[\textit{inter alia}]{ni2009mining,plamadua2013mining,schwenk2021wikimatrix}. We add to the body of work that exploits Wikipedia as a massively multilingual text resource by using it to build pretraining data for HMDE. Concretely, for a set of languages $L = \{L_1, L_2, \dots, L_n \}$, we first fetch monolingual portions from the Wiki-40B corpus.\footnote{Available in Tensorflow datasets: \url{https://www.tensorflow.org/datasets/catalog/wikipedia}} We then identify articles in different languages that are about the same concept (via the \texttt{wikidata\_id} field) and keep only those concepts for which pages are found in at least two languages from $L$. For each such concept with pages $p_1, p_2, \dots, p_m$ in $m$ different languages, we create all possible cross-lingual pairs of articles ($p_i$, $p_j$) covering the same concept. For each pair ($p_i$, $p_j$), we then leverage Wikipedia metadata -- namely mapping of Wikipedia pages into its hierarchy of categories -- to select an article $n_i$ from the same monolingual Wikipedia as $p_i$ (i.e., written in the same language as $p_i$) that belongs to (at least one) same Wikipedia category as $p_i$. This yields triples ($p_i$, $p_j$, $n_i$) from which we create cross-lingual positives ($p_i$, $p_j$) and their corresponding monolingual hard negatives ($p_i$, $n_i$) for our contrastive objective (see \S\ref{sec:objective}).

On the one hand, the quality of MMTs' representations of a particular language depends on the size of the pretraining corpora of that language \cite{hu2020xtreme,lauscher2020zero}. On the other hand, multilingual model training with instances from linguistically diverse languages may generalize better to unseen languages \cite{chen2019multi,ansell2021mad}. Most resourced languages, however, tend to be Indo-European \cite{joshi2020state}, putting corpus size and linguistic diversity at odds. We thus create two different datasets, each emphasis one of these two aspects: (1) XLW-4L is built starting from four high-resource Indo-European languages: English, German, French, and Italian; (12) XLW-12L is built starting from a set of 12 linguistically diverse languages: English, French, Russian, Japanese, Chinese, Hungarian, Finnish, Arabic, Persian, Turkish, Greek, and Malay. With 1.1M triples ($p_i$, $p_j$, $n_i$), XLW-4L is almost twice as large as XLW-12L (which encompasses 592K triples), despite encompassing three times fewer languages: this is primarily because there are many more shared concepts between large Wikipedias of XLW-4L (e.g., German and Italian) than between smaller Wikipedias of XLW-12L (e.g., Turkish and Malay).\footnote{Per-language statistics of the datasets are in the Appendix.}

\subsection{Evaluation Tasks and Datasets}
\label{sec:downstream}

HMDE is meant to be a general-purpose multilingual document encoder. It thus needs to be useful both (1) when fine-tuned for a supervised document-level task, and (2) as a standalone document encoder. We thus evaluate HMDE in (1) zero-shot cross-lingual transfer for supervised document classification (XLDC) and (2) unsupervised cross-lingual document retrieval (CLIR).   

\paragraph{XLDOC.} Regular MMTs (e.g., mBERT or XLM-R) are primarily used in zero-shot cross-lingual transfer for supervised NLP tasks: an MMT fine-tuned on task-specific training data in a resource-rich language is used to make predictions for language(s) without task data. We evaluate HMDE in exactly the same zero-shot cross-lingual transfer setup, only for a document-level task -- topical document classification. We fine-tune HMDE in the standard manner, by stacking a softmax classifier on top the output of the document-level encoder. With $\mathbf{d}$ as HDME's encoding of the input document $d$, classifier's prediction is computed as: 
\begin{equation}
 \mathbf{y} = \mathit{softmax}\left(\mathbf{W}\cdot\mathbf{d} + \mathbf{b}\right)
\end{equation}
\noindent with $\mathbf{W} \in \mathbb{R}^{C \times h}$ and $\mathbf{b} \in \mathbb{R}^{C}$ as classifier's trainable parameters (and $C$ as the number of classes).  

We fine-tune HMDE on the English training portion of the MLDOC dataset \cite{SCHWENK18.658} and evaluate its performance on the test portions of all other (target) languages. MLDOC is a subset of the Reuters Corpus Volume 2 (RCV2), with training, development, and test portions in 8 languages (English, Spanish, German, French, Italian, Russian, Japanese and Chinese), consisting of 1000, 1000, and 4000 documents, respectively. News stories are categorized into $C = 4$ semantically closely related classes (\textit{Corporate/Industrial}, \textit{Economics}, \textit{Government/Social}, and \textit{Markets}).     


\paragraph{CLIR.} We evaluate the effectiveness of HMDE as a standalone document encoder in an unsupervised cross-lingual document retrieval task: queries (short text) in one language are fired against a collection of documents written in another language. We adopt a simple retrieval model: we rank documents in decreasing order of cosine similarity of their embeddings $\mathbf{d}$, produced by the HMDE, with the embedding $\mathbf{q}$ of the query, $\cos(\mathbf{d}, \mathbf{q})$. We obtain the query embedding $\mathbf{q}$ by encoding the query only with HMDE's lower (sentence-level) transformer: $\mathbf{q}$ is the transformed representation of the beginning-of-sequence ([BOS]) token.     

We carry out the evaluation on CLEF-2003,\footnote{\url{http://catalog.elra.info/en-us/repository/browse/ELRA-E0008/}} a popular CLIR benchmark, including the following languages: English (EN), German (DE), Italian (IT), Finnish (FI) and Russian (RU). Following prior work \cite{glavavs2019properly,litschko2022cross}, we evaluate HMDE on 9 language pairs (with first language being the query language): EN-{FI, DE, IT, RU}, DE-{FI, IT, RU}, FI-{IT, RU}. For each language pair we work with 60 queries and document collections of following sizes: RU -- 17K, FI -- 55K, IT -- 158K, and DE -- 295K. 



\subsection{Baseline Models}
\label{sec:baselines}

There are two main alternatives to hierarhical (long) document encoding. The first is to (i) fragment the document into smaller segments, (ii) encode each segment with a regular pretrained MMT (e.g., vanilla MMT like XLM-R or a multilingual sentence encoder like LaBSE), and (iii) aggregate the document representation from the embeddings of segments. The second is to train a multilingual sparse-attention encoder, akin to \cite{Sagen1545786}.    

\paragraph{MMT with a Sliding Window (LaBSE-Seg).} For fair comparison, we use LaBSE \cite{feng2022language} -- the same pretrained MMT that we use for the initialization of the lower transformer in HMDE -- to independently encode overlapping segments of the input document. We break down the document into segments of length $N_S$ tokens. Following \newcite{dai2022revisiting}, who find that overlapping segments alleviate the context fragmentation problem, we make adjacent segments overlap in $N_S/3$ tokens. After encoding each segment with LaBSE, we average-pool the document representation \textbf{d} from the set of segment embeddings. In XLDX (topical document classification) this average of segment embeddings is fed into the classification head. In CLIR, it is compared with the LaBSE encoding of the query.                  

\paragraph{Multilingual Longformer (mLongformer).} Longformer architecture \cite{beltagy2020longformer} combines local-window attention with global attention, resulting in a hybrid attention mechanism, the memory requirements of which scale linearly with the input length. \newcite{beltagy2020longformer} additionally propose multi-step procedure for initializing Longformer's parameters based on the parameters of a pretrained regular transformer (e.g., in the case of monolingual English Longformer from RoBERTa \cite{liu2019roberta}) and then further train the Longformer via masked language modeling (MLM). We train the multilingual Longformer following the same procedure: for fair comparison with HMDE, we initialize its parameters from the parameters of LaBSE and carry out the additional MLM training on XLW-4L, the same corpus on which we train HMDE.

\section{Results and Discussion}
\label{sec:res}

\setlength{\tabcolsep}{8pt}
\begin{table*}[t!]
    \centering
    \normalsize
    \begin{tabular}{lc cc cc cc cc}
    \toprule
    Model & En & Es & De & Fr & It & Ru & Ja & Zh & AVG \\ 
\midrule \multicolumn{9}{l}{\textit{Standard Multilingual Transformers}} \\ \midrule
LaBSE & 95.5 & 79.0 & 89.6 & 87.2 & 76.8 & 63.9 & \textbf{80.8} & 86.1 & 82.4 \\ 
XLM-R \citep{dong-etal-mldocbaseline} & 93.0 & 84.6 & \textbf{92.5} & 87.1 & 73.2 & 68.9 & 78.2 & 85.8 & 83.0  \\
mBERT \citep{zhao-etal-2021-closer} & \textbf{96.9} & 81.9 & 88.3 & 83.1 & 74.1 & 72.3 & 74.6 & 84.4 & 82.0  \\ 
\midrule \multicolumn{9}{l}{\textit{Multilingual Long Document Encoders}} \\ \midrule
LaBSE-Seg & 94.0 & 82.9 & 90.2 & 89.9 & 78.1 & 71.9 & 75.5 & 88.4 & 84.0  \\
mLongformer (XLW-4L) & 95.8 & \textbf{87.0} & 93.4 & 91.9 & \textbf{80.6} & 71.7 & 79.5 & 88.5 & 86.1 \\
HMDE (XLW-4L) & 95.4 & 85.6 & 91.2 & \textbf{92.0} & 78.5 & \textbf{83.9}	& 76.3 & \textbf{89.5} & \textbf{86.8} \\
\bottomrule
    \end{tabular}
    \caption{Performance of HDME compared against standard MMTs and baseline multilingual long-document encoders on supervised topical document classification (MLDOC). Performance (except En) for zero-shot cross-lingual transfer: all models are fine-tuned only on English training data. \textbf{Bold}: best performance in each column.}
    \label{tab:mldoc_results}
\end{table*}

\setlength{\tabcolsep}{2.4pt}
\begin{table*}[t!]
    \centering
    \normalsize
    \begin{tabular}{lccc cccc ccc} \toprule
    Model & En–Fi & En–It & En–Ru & En–De & De–Fi & De–It & De–Ru & Fi–It & Fi–Ru & AVG \\ 
\midrule \multicolumn{9}{l}{\textit{Standard Multilingual Transformers}} \\ \midrule
LaBSE & .247 & .224 & .131 & .138 & .247 & .214 & .135 & .211 & .125 & .186 \\
mBERT\,\cite{litschko2022cross}& .145 & .146 & \textbf{.167} & .107 & .151 & .116 & \textbf{.149} & .117 & .128 & .136 \\
\midrule \multicolumn{9}{l}{\textit{Multilingual Long Document Encoders}} \\ \midrule
LaBSE-Seg & .243 & .169 & .107 & .194 & .268 & .178 & .104 & .153 & .014 & .159 \\  
mLongformer (XLW-4L) & .150 & .088 & .094 & .082 & .190 & .072 & .120 & .097 & .091 & .109 \\ 
HMDE (XLW-4L) & \textbf{.380} & \textbf{.282} & .141 & \textbf{.326} &	\textbf{.352} & \textbf{.259} & .130 & \textbf{.238} & \textbf{.129} & \textbf{.249} \\
\bottomrule
    \end{tabular}
    \caption{Performance of HDME compared against standard MMTs and baseline multilingual long-document encoders on unsupervised cross-lingual document retrieval (CLEF-2003). \textbf{Bold}: best performance in each column.}
    \label{tab:clir_results}
\end{table*}


We first report and discuss the main results we obtain with HMDE on XLDC and CLIR (in \S\ref{sec:mainres}). In a series of follow-up experiments, we further analyze key design choices for HMDE (\S\ref{sec:analysis}). 

\subsection{Main Results}
\label{sec:mainres}

\paragraph{Cross-lingual Document Classification.} Table \ref{tab:mldoc_results} compares HMDE trained on XLW-4L against several standard and long document multilingual encoders: besides the baselines introduced in \S\ref{sec:baselines}, for completeness we add the results for vanilla LaBSE (i.e., without sliding over the long document) and models based on XLM-R and mBERT reported by \newcite{dong-etal-mldocbaseline} and \newcite{zhao-etal-2021-closer}, respectively.   
Expectedly, all long-document encoders outperform all of the standard MMTs. mLongformer and HMDE generally exhibit similar performance, surpassing the performance of segmentation-based LaBSE-Seg for virtually all languages. Comparable performance of mLongformer and HMDE suggests that in the presence of task-specific fine-tuning data it does not really matter whether we aggregate document representations in a flat or hieratrchical fashion. What is particularly encouraging is that both HDME and mLongformer exhibit strong performance for languages that they did not observe in document-level pretraining: Spanish, Russian, Japanese, and Chinese.\footnote{LaBSE, with whose parameters both HMDE and mLongofrmer were initialized before document-level pretraining, however, was exposed to all of these languages in its own sentence-level pretraining.}\textsuperscript{,}\footnote{Performance across languages \textit{not} directly comparable as MLDOC test sets are not parallel across languages.}    


\paragraph{Cross-lingual Retrieval.} The results for unsupervised CLIR are shown in Table \ref{tab:clir_results}. Like in XLDC, we additionally report the results for LaBSE that encodes only the beginning of the document (without sliding) as well as for mBERT, reported by \newcite{litschko2022cross}. CLIR, in which multilingual transformers are used as standalone document encoders without any task-specific fine-tuning, tell a very different story from supervised XLDC results. HMDE drastically outperforms mLongformer, indicating that, much like the vanilla MMTs, mLongformer requires fine-tuning and cannot encode reliably encode documents ``out of the box''. HMDE also substantially outperforms LaBSE-Seg, the long-document encoder based on sliding LaBSE over the document. Interestingly, vanilla LaBSE, which encodes only the beginning of the document, also outperforms its sliding counterpart LaBSE-Seg, which is exposed to the entire document. We believe that this is because (1) in CLEF, retrieval-relevant information often occurs at the beginnings of documents and in such cases (2) LaBSE-Seg's average-pooling over all document segments then dilutes the encoding of query-relevant content. Importantly, HMDE in CLIR also seems to generalize very well to languages unseen in its document-level pretraining (in particular for Finnish documents).


\subsection{Further Analysis}
\label{sec:analysis}

We next empirically examine how different choices in HDME's design and pretraining affect its performance, focusing on: (i) linguistic diversity and size of the pretraining corpus (XLW-4L vs. XLW-12L), (ii) freezing of the lower transformer (i.e., LaBSE weights) after initialization, and (iii) initializing it with the weights of XLM-R as the standard MMT (vs.\,initialization with LaBSE as the sentence encoder). We provide a further ablations on document segmentation (sentences vs.\,token sequences ignorant of sentence boundaries) in the Appendix \ref{sec:ablations}.        

\begin{figure}[t!]
    \centering
    \includegraphics[width=\linewidth]{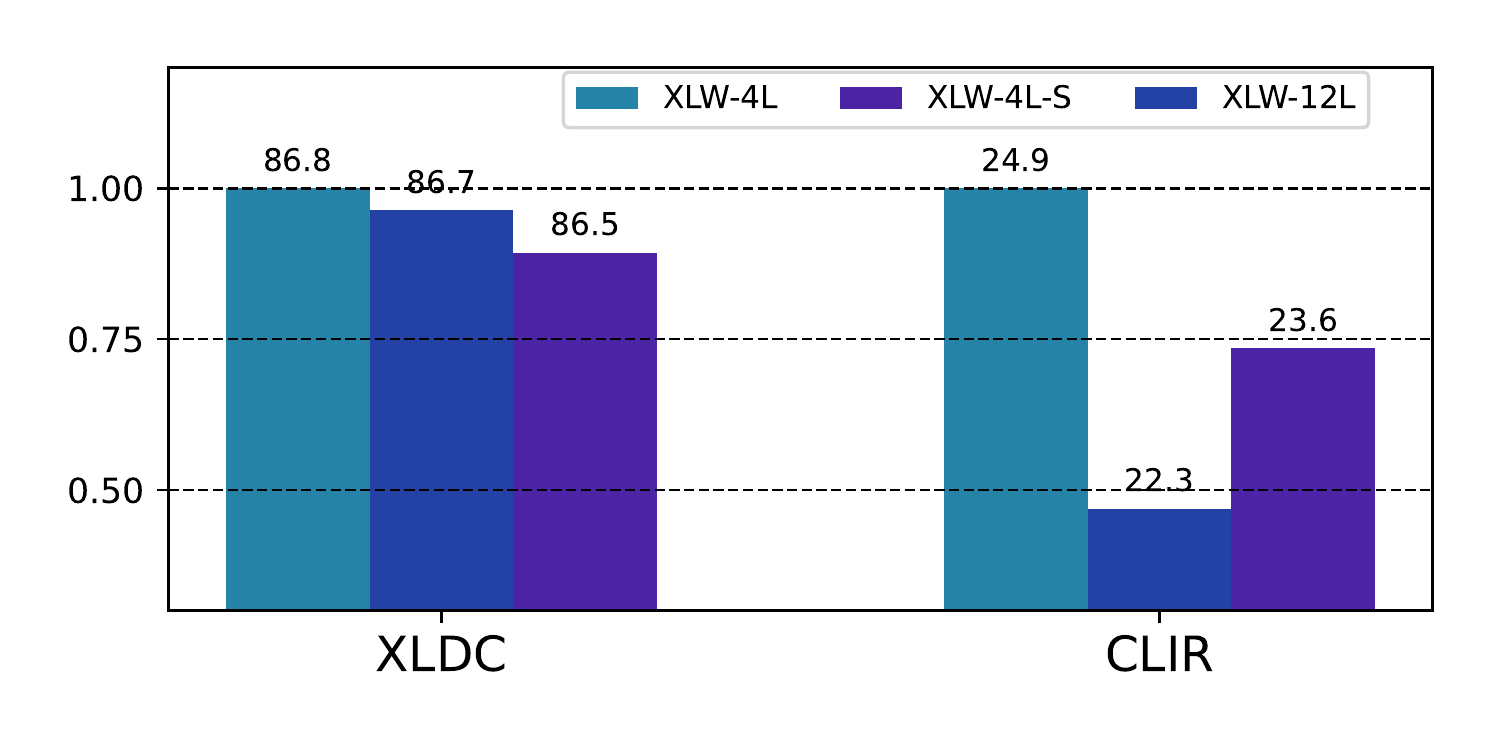}
    \caption{Performance of HMDE when pretrained on different datasets. Results are averages across all test languages (XLDC) and language pairs (CLIR).}
    \label{fig:sizeld}
\end{figure}

\paragraph{Pretraining Data: Linguistic Diversity vs.\,Size.} As discussed in \S\ref{sec:data}, we prepare two different corpora for HMDE pretraining: XLW-4L, which is larger (1.1M instances) but encompasses only four major Indo-European languages and XLW-12L, which is smaller (590K instances) but has documents from a set of 12 linguistically diverse languages. To control for the size, and assess the effect of linguistic diversity alone, we randomly downsample XLW-4L, creating a 4-language dataset XLW-4L-S that matches in size XLW-12L. Figure \ref{fig:sizeld} shows the downstream performance of HMDE when pretrained on each of these three datasets. 

Comparison between XLW-4L and XLW-4L-S (same languages, different dataset size) shows that our flavor of cross-lingual contrastive pretraining (\S\ref{sec:objective}) leads to a fairly sample-efficient pre-training: cutting the training data almost in half leads to small performance drops (mere 0.3 accuracy points in XLDC; 1.3 MAP points in CLIR). Comparison between XLW-4L-S and XLW-12L (same size, different language sets) quantifies the role of linguistic diversity in pretraining. Somewhat surprisingly, the more linguistically diverse pretraining on XLW-12L does not bring better performance compared to ``Indo-European-only'' pretraining on XLW-4L-S: while they perform comparably on XLDC, more diverse pretraining (XLW-12L) leads to worse CLIR performance (-1.3 MAP points on average). We hypothesize that this is due to higher-quality of representation of the four Indo-European languages (EN, DE, FR, IT) in LaBSE (owing to their overrepresentation in LaBSE's pretraining), with which we initialize the lower transformer of HMDE. We find this result to be particularly encouraging, as -- together with the observation that HMDE generalizes well to languages unseen in its document-level pretraining -- it suggests that document-level pretraining itself does not necessarily need to be massively multilingual in order to yield successful massively multilingual document encoders.   

\paragraph{Lower Transformer.} We next investigate two aspects of the lower-transformer: (1) with which weights to initialize it and (2) whether it pays off to update its parameters during the document-level pretraining. For the former, we compare our default LaBSE-based initialization (with LaBSE as a sentence-specialized multilingual encoder) against the initialization with weights of XLM-R, as the vanilla multilingual MMT. To answer the latter, we additionally train HMDE by freezing its lower transformer in document-level pretraining. Table \ref{tab:ablations} summarizes the results of these ablations.       

\setlength{\tabcolsep}{6.5pt}
\begin{table}[t!]
    \normalsize
    \centering
    \begin{tabular}{l l c c}
    \toprule
    Model & Updates & XLDC & CLIR \\ \midrule
    HMDE-LaBSE & \textit{Updated} & \textbf{86.8} & \textbf{0.249} \\
    \midrule
    HMDE-LaBSE & \textit{Frozen} & 85.9 & 0.167 \\
    HMDE-XLM-R & \textit{Updated} & 83.9 & 0.135 \\
    \bottomrule
    \end{tabular}
    \caption{HMDE results for different choices w.r.t. to initialization and training of the lower transformer. Training for all three variants carried out on XLW-4L. Results are averages across all test languages (XLDC) and language pairs (CLIR).}
    \label{tab:ablations}
\end{table}

While freezing the lower transformer after initialization leads to much faster training, it results in poorer document encoder, especially if used for standalone document encoding, without task-specific fine-tuning\footnote{The parameters of the lower-transformer are always updated in XLDC fine-tuning, even if we froze them in document-level pretraining.} (HMDE-LaBSE \textit{Updated} vs. \textit{Frozen}; 1 accuracy point drop in XLDC vs. 8 MAP points drop in CLIR). Initializing HDME's lower transformer with LaBSE weights leads to much better downstream performance compared to initialization with XLM-R which is not specialized for sentence-level semantics.

\section{Related Work}

We position our contributions w.r.t. three related lines of work: (1) pretraining long-document encoders, (2) self-supervised pretraining for retrieval, and (3) mining parallel documents.            

\paragraph{Long-Document Encoders.} 
Hierarchical \cite{zhang-etal-2019-hibert,yang2020beyond,glavas2020two} and sparse-attention-based encoders \cite{beltagy2020longformer,zaheer2020big,tay2020sparse} already discussed in \S\ref{sec:introduction} account for the vast majority of long-document encoding approaches. \newcite{dai2022revisiting} extensively compare Longformer \cite{beltagy2020longformer} against hierarchical transformers on various long-document classification tasks, showing that the latter exhibit slightly better performance, especially if the lower encoder encodes overlapping segments. 
\newcite{ding-etal-2021-ernie} propose a different, segmentation-based model based on recurrence transformers \cite{dai2019transformer}, designed to remedy for context fragmentation with a retrospective feed mechanism: each segment is encoded twice -- after initial left-to-right segment with a recurrent transformer, segment representations are further mutually contextualized bidirectionally. Their training couples MLM-ing with a segment reordering objective. 

The vast majority of work on pretraining encoders for long documents focuses on monolingual (mainly English) models. The few multilingual exceptions \cite{yu2021cross,Sagen1545786} derive a multilingual Longformer from standard MMTs (XLM-R and mBERT) in exactly the same fashion in which the original work \cite{beltagy2020longformer} pretrains English Longformer after initialization from RoBERTa weights. In this work, we replicated this effort, evaluating mLongformer as the main baseline for HMDE.


\paragraph{Pretraining for Retrieval.}

Self-supervised and distantly-supervised approaches have recently been proposed for pretraining documents encoders specifically for the task of document retrieval \cite{izacard2022unsupervised,yu2021cross,gao2022precise}. \newcite{izacard2022unsupervised} pretrain Contriever -- a BERT-based document encoder with an objective based on the inverse cloze task \cite{lee2019latent}: a positive query-document pair is created by extracting a span of text from the document and using it as a ``query''; they train with a contrastive objective that scores the document from which the query was extracted higher than other documents. \newcite{gao2022precise} feed queries as prompts to a generative language model, which then generates document; they then use Contriever to embed this synthetic document and find most similar real documents in the collection, finally 
fine-tuning Contriever on query-document pairs obtained this way. 
In a manner similar to ours, \newcite{yu2021cross} leverage Wikipedia as a source of quasi-parallel data: while we exploit document-level alignments, they leverage section-level aligments to create positive cross-lingual training instances for paragraph retrieval: a section title (``query'') in one language is coupled with the section body (``document'') in another language; they then train a multilingual Longformer initialized from mBERT with a combination of query MLM-ing and contrastive relevance ranking. In contrast to these efforts, we create a general-purpose (i.e., task-agnostic) multilingual document encoder that can both be fine-tuned for supervised tasks and used as a standalone document embedder. 

\paragraph{Mining Parallel Documents.}

Mining parallel documents -- a task which aims to identify mutual translations in a large document collection and is often used as a first step in extracting parallel sentences \cite[\textit{inter alia}]{resnik2003web,uszkoreit2010large,schwenk-2018-filtering} -- is the task that bears most resemblance to our pretraining. Transformer-based approaches to the task \cite{guo2019hierarchical,el2020massively,gong2021lawdr} typically aggregate document-level representations from multilingual sentence embeddings. The work of \newcite{guo2019hierarchical} is arguably most related to ours: they train a hierarchical encoder with a simple feed-forward net as the upper encoder that independently transforms precomputed sentence embeddings: document embedding is then the average of feed-forward-transformed sentence embeddings. 
The model is trained bilingually (English-Spanish and English-French) with a contrastive objective on a huge silver-standard corpus of parallel documents (13M and 6M document pairs, respectively) and evaluated on the very same task of parallel document mining. Our work differs in two crucial aspects: (1) while \cite{guo2019hierarchical} train \textit{bilingual} models for recognizing parallel documents,  we train a single general-purpose massively multilingual document encoder; (2) we train on a much smaller corpus of comparable (not parallel) documents, readily available from Wikipedia. Both aspects make HMDE much more widely applicable, for both supervised and unsupervised document-level tasks and any of the languages from LaBSE's pretraining (as HMDE's lower encoder is initialized with LaBSE's weights).                   

\section{Conclusion}

In this work, we pretrain a multilingual document encoder based on a hierarchical transformer architecture (HMDE), and initialize its lower-level encoder with the weights of a state-of-the-art multilingual sentence encoder. We leverage Wikipedia as a rich source of quasi-parallel long documents and train HDME with a contrastive cross-lingual document matching objective. We show that the obtained model is a general-purpose multilingual document encoder that can successfully be both (1) fine-tuned for document-level cross-lingual transfer and (2) used as a document embedding model out of the box. Our results render HMDE substantially more effective than both multilingual Longformer and segmentation-based document encoding. Crucially, HMDE generalizes well to languages unseen in its document-level pretraining. Our follow-up experiments reveal that the size of the pretraining corpus affects the performance more than the number and diversity of languages involved, suggesting that reliable massively multilingual document encoders do not necessarily require equally massively multilingual pretraining.

\section*{Limitations}

Because we initialize the lower transformer of HMDE with LaBSE \cite{feng2022language}, the set of languages that HMDE ``supports'' out of the box is bound to the set of 109 languages included in LaBSE's pretraining.\footnote{The full list is provided in Table 10 of the Appendix in \cite{feng2022language}.} This means that HMDE will, in principle, be less effective as a document encoder for other languages.\footnote{Not necessarily the case only for unseen that are close relatives to some of the high-resource languages seen in LaBSE's pretraining.} HDME, like LaBSE, should in principle be useless for languages written in a script that LaBSE (or in fact, mBERT, from which LaBSE borrows the vocabulary and pretrained subword embeddings) has not seen in its pretraining, as the corresponding tokenizer will produce a sequence of unknown tokens ([UNK]). This means that HMDE, much like the rest of existing multilingual encoders, supports only a small fraction of world's 7000+ languages \cite{joshi2020state}. Moreover, all languages included in our evaluation datasets -- MLDOC and CLEF -- are covered by this set of 109 languages, which means that the average performance we report is likely a gross overestimate for languages unseen in LaBSE's pretraining. Further, HMDE leverages Wikipedia for training (with sets of either 4 or 12 languages, see \ref{sec:data}) -- the number of Wikipedia pages (and more generally, digital footprint of a language on the web) varies tremendously across languages, effectively limiting the selection of languages for HMDE's document-level pretraining. Our results (see \ref{sec:mainres}), however, show that HMDE generalizes well to languages not seen in its document-level pretraining.            

Further, HMDE is  implemented as a Bi-Encoder (aka Siamese network), which means that for a given pair of documents in a training example (positive or negative pair), it separately encodes each of the documents. Cross-Encoder architecture, in which the documents would be concatenated before encoding, would have the advantage of allowing the encoder to contextualize the token/sentence representations of one document with those of the other before the computation of their similarity score. Cross-encoding architectures have been shown effective, albeit not efficient (i.e., slow) in training for document retrieval, in which the (short) query is concatenated with the (long) document \cite{macavaney2020teaching,shi-etal-2020-cross,rosa2022defense}. We do not explore cross-encoding in our work; in our case, it implies joint encoding of the concatenation of two long documents (in different languages), arguably exploding in GPU memory occupancy and possibly preventing us from fitting even single-instance batches on our GPU cards.

\section*{Ethical Considerations} 
We do not test HMDE explicitly to check whether the representations it produces reflect negative societal biases and stereotypes (e.g., sexism or racism), but given that its lower encoder is initialized from LaBSE's weights, it would not be surprising if this was the case. If so, many of the existing techniques from the literature designed to debias pretrained language models \cite{qian2019reducing,barikeri2021redditbias,guo2022auto} could be applied to HMDE too, and in principle ``as-is'' (i.e., without special modifications).

\bibliography{references}
\bibliographystyle{acl_natbib}

\clearpage
\appendix
\section{Appendix}
\label{sec:appendix}

\subsection{Training and Optimization Details}
\label{sec:training}

In all training procedures, we use AdamW \cite{loshchilov2019decoupled} as the optimization algorithm. 

\paragraph{HMDE Pretraining.} We set the maximal sentence length for HMDE, input to its lower-level transformer (initialized with LaBSE weights) to $128$ tokens. For fair comparison, we set the segment size of the LasBSE-Seg baseline also to $N_S = 128$ tokens. For fair comparison against mLongformer, we limit the maximal document length for HMDE to $32$ sentences, not to exceed the mLongformer's maximal input length of $4,096$ tokens. In our main set of experiments, the document-level (upper) transformer consists of 2 transformer layers, with GELU activation \cite{hendrycks2016gaussian}, layer normalization ($\epsilon = 1\mathrm{e}^{-12}$), and feed-forward sublayer with hidden size of $2048$. The dropout rate for the upper transformer is set to $0.1$. We train in batches of size $N = 2$ with the gradient accumulation over $64$ batches for $1$ full epoch,\footnote{Note that batch size $N = 2$ in our contrastive objective (see \S\ref{sec:objective}) implies only one in-batch negative pair (besides the hard negative) for each positive pair.} with the initial learning rate of $1\mathrm{e}^{-5}$, linear scheduling and 1000 warm-up steps. 
   
%

            

\paragraph{mLongformer Pretraining.}
We train the mLongformer model (also initialized from LaBSE), also for $1$ full epoch  via MLM-ing, masking out 15\% of tokens. We train with the initial learning rate of $1\mathrm{e}^{-5}$ with weight decay of $0.01$ and 500 warm-up steps. We train in batches of size 2, accumulating gradients over 32 batches.     

\paragraph{XLDC Fine-Tuning.} We fine-tune both HMDE and mLongformer for topical document classification with the learning rate of $2\mathrm{e}^{-5}$ and without weight decay (with a 200 warm-up steps). We train in batches of size $4$ for $50$ epochs, accumulating gradients over $8$ batches. Model selection was carried out based on the performance on the English validation portion of the MLDOC dataset, with early stopping if validation loss did not improve over $7$ epochs.


\subsection{Additional Ablation}
\label{sec:ablations}

\setlength{\tabcolsep}{4.5pt}
\begin{table}[t!]
    \normalsize
    \centering
    \begin{tabular}{l l c c}
    \toprule
    Model & Segmentation & XLDC & CLIR \\ \midrule
    HMDE-LaBSE & \textit{Sentence} & 86.8 & 0.249 \\
    HMDE-LaBSE & \textit{Chunk} & 85.4 & 0.224 \\
    \bottomrule
    \end{tabular}
    \caption{HMDE results for different choices w.r.t. to document segmentation. Training for both variants carried out on XLW-4L. Results are averages across all test languages (XLDC) and language pairs (CLIR).}
    \label{tab:chunks}
\end{table}

We additionally test our design decision to segment the document into sentences, and encode sentences with the lower-level transformer (the weights of which are initialized from LaBSE). To this end, we compare our default strategy of segmenting input documents into sentences against a less-informed segmentation into consecutive chunks of $128$ tokens. Table \ref{tab:chunks} shows the results of this comparison. Unsurprisingly -- given that the lower encoder is initialized with the weights of a pretrained \textit{sentence} encoder -- sentence-based segmentation is more effective, although chunking does not trail by much. 

\end{document}